\documentclass{article}
\usepackage{spconf,amsmath,graphicx}

\usepackage{url}
\usepackage{cite}
\usepackage[hidelinks]{hyperref}
\usepackage{booktabs}
\usepackage{multirow}
\usepackage{paralist}
\title{What do self-supervised speech and speaker models learn?\\New findings from a cross model layer-wise analysis}

\name{Takanori Ashihara, Marc Delcroix, Takafumi Moriya, Kohei Matsuura, Taichi Asami, Yusuke Ijima}
\address{NTT Corporation, Japan}

\begin{document}
\ninept

\maketitle

\begin{abstract}
\vspace{-0.1cm}
Self-supervised learning (SSL) has attracted increased attention for learning meaningful speech representations.
Speech SSL models, such as WavLM, employ masked prediction training to encode general-purpose representations.
In contrast, speaker SSL models, exemplified by DINO-based models, adopt utterance-level training objectives primarily for speaker representation.
Understanding how these models represent information is essential for refining model efficiency and effectiveness.
Unlike the various analyses of speech SSL, there has been limited investigation into what information speaker SSL captures and how its representation differs from speech SSL or other fully-supervised speaker models.
This paper addresses these fundamental questions.
We explore the capacity to capture various speech properties by applying SUPERB evaluation probing tasks to speech and speaker SSL models.
We also examine which layers are predominantly utilized for each task to identify differences in how speech is represented.
Furthermore, we conduct direct comparisons to measure the similarities between layers within and across models.
Our analysis unveils that 1) the capacity to represent content information is somewhat unrelated to enhanced speaker representation, 2) specific layers of speech SSL models would be partly specialized in capturing linguistic information, and 3) speaker SSL models tend to disregard linguistic information but exhibit more sophisticated speaker representation.
\end{abstract}

\vspace{-0.1cm}

\vspace{-0.3cm}
\section{Introduction}
\label{sec:intro}
\vspace{-0.3cm}
Self-supervised learning (SSL) for speech representation has been a prominent technique for leveraging unsupervised data without any hand-crafted labels~\cite{ssl_review}.
Recent speech SSL studies~\cite{wav2vec2,hubert,wavlm} have successfully demonstrated superior performance in diverse tasks~\cite{superb, speechglue} by encoding general-purpose representations such as speaker and linguistic knowledge from speech data alone.
To capture diverse representations that disentangle speech variability hierarchically, speech SSL methods typically employ deep network architectures with substantial model capacity~(e.g.,~12--24 transformer blocks with almost 94M--315M parameters).
Additionally, these methods incorporate pretext tasks that consider the sequential nature of speech, including masked prediction tasks~\cite{hubert,wavlm} and future prediction tasks~\cite{cpc,apc}, inspired by the natural language processing community.
\par
In concurrence with the impressive progress of speech SSL, SSL for acquiring mainly speaker representations, hereinafter referred to as speaker SSL, has also been a flourishing research area~\cite{spkssl_mse,spkssl_simclr_moco,spkssl_vicreg,spkssl_dino_er,spkssl_c3dino,spkssl_dino_compara,spkssl_ca_dino,spkssl_rdino}.
While the pretext tasks for speech SSL are designed to discover contextualized frame-level representations, speaker SSL specializes in capturing utterance-level information such as speaker and emotion knowledge.
Thereby, in speaker SSL, the usual training target is a single pseudo-label assigned to an input segment/utterance.
Consequently, the output of the model comprises utterance-wise embeddings, which can be directly utilized for tasks,~e.g.,~automatic speaker verification~(ASV) without the need for additional supervised fine-tuning.
In contrast to speech SSL, speaker SSL aims to encode representations for specific purposes, and hence, employs lightweight speaker models such as ECAPA-TDNN~\cite{ecapa_tdnn} having considerably fewer parameters~(typically 6M--15M).
\par
Despite the progress of speaker SSL, there has been little understanding of what entangled property in spoken utterances is captured by these models, as opposed to the several analytical studies conducted on speech SSL representation~\cite{sphssl_lwise_anal,speechglue,sphssl_similarity,sphssl_sa_anal,sphssl_prob_anal,ashi_deep_wide,anal_mpc}.
More concretely, since the representation from speaker to linguistic factor is presumed to be distributed from shallow to deep layers in speech SSL, do speaker SSL models capture either a similar distribution or opposite?
In addition, as speaker SSL models have lagged behind in the ASV performance compared with fully-supervised speaker models~\cite{spkssl_dino_compara}, we intend to elucidate how these two representations differ.
\par
In this paper, to explore the representation differences between speech SSL, speaker SSL and fully-supervised speaker models, we apply two analysis frameworks to these models.
The first involves using diverse speech-processing probing tasks to compare the capacity of these models to capture speech properties by analyzing their performance trends.
Following SUPERB~\cite{superb}, the downstream model takes as input a weighted sum of the layers from the upstream models.
Thus, learnable weights provide insights into the importance of each layer in addressing the tasks.
The second analysis, which complements the above indirect approach, involves a direct comparison that measures the similarity of representations in each hidden layer, both within the same model and between different models~\cite{lincka}.
The main findings of our analysis are:
\begin{inparaenum}[(1)]
\item the capacity to accurately represent non-speaker information such as spoken content and semantics is poorly correlated with enhanced speaker representation,
\item both speaker SSL and fully-supervised speaker models disentangle phoneme information in early layers in contrast to speech SSL models, and
\item speech SSL models exhibit a characteristic pattern in their deeper layers, seemingly associated with linguistic information partly, while the representation of these deeper layers is largely lacking in both speaker models.
\end{inparaenum}
We hope that the result of this analysis will drive the development of more efficient networks and effective training procedures such as the previous work~\cite{sphssl_lwise_anal}.

\vspace{-0.3cm}
\section{Related work}
\vspace{-0.3cm}
Some previous works have been dedicated to analyzing speech SSL representation~\cite{sphssl_lwise_anal,speechglue,sphssl_similarity,sphssl_sa_anal,ashi_deep_wide,sphssl_prob_anal,anal_mpc}.
For example,~\cite{sphssl_lwise_anal} offered a layer-wise analysis to elucidate how the representations of acoustic, phonetic, and word-level factors were encoded across layers.
Additionally, \cite{speechglue} provided a benchmark for exploring more high-level linguistic representations (e.g.,~lexicon, semantics, and syntax) captured by speech SSL.
\cite{sphssl_similarity} compared the similarity of only last-layer output between SSL methods/architectures.
However, to the best of our knowledge, there is no comprehensive research comparing layer-wise representations within and between speech SSL, speaker SSL, and fully-supervised speaker models.

\vspace{-0.2cm}
\section{Method}
\label{sec:method}
\vspace{-0.2cm}

\subsection{Training methods for speaker models}
\label{ssec:met:spk_rep}
\begin{figure}[t]
  \centering\hspace*{-1cm}
  \includegraphics[width=6cm]{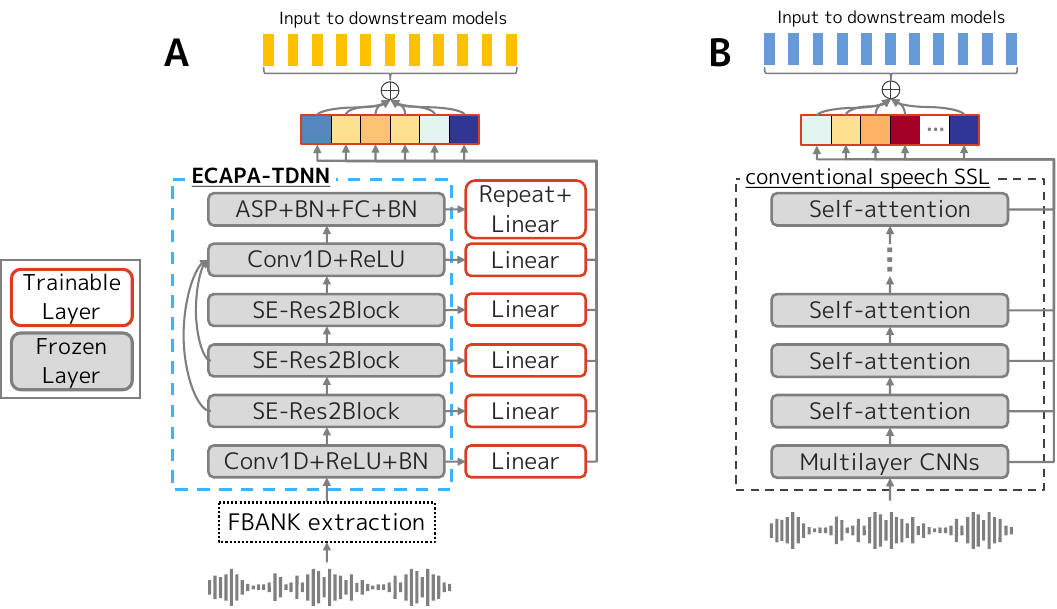}
  \vspace{-0.3cm}
  \caption{Schematic diagrams of benchmarking (A) ECAPA-TDNN and (B) conventional speech SSL upstream models with several self-attention blocks on SUPERB~\cite{superb}.}
  \label{fig:ecapa_s3prl}
  \vspace{-0.3cm}
\end{figure}

\vspace{-0.1cm}
One conventional approach for acquiring speaker embeddings is to train ECAPA-TDNN~\cite{ecapa_tdnn} in a supervised fashion.
This model is widely used in speaker-related tasks.
Fig.~\ref{fig:ecapa_s3prl}A shows ECAPA-TDNN.
This model consists of five frame-level blocks~(i.e.,~a standard 1D convolution layer, multiple Squeeze-Excitation(SE)-Res2Blocks, and multilayer feature aggregation followed by a 1D convolution layer) and a segment-level block~(i.e.,~an attentive statistics pooling layer followed by a fully-connected layer as a bottleneck layer).
In our experiment, to build a supervised speaker model, ECAPA-TDNN is optimized through a speaker identification (SID) task with angular additive margin softmax~(AAM-Softmax) loss followed by large margin fine-tuning, which is a widely accepted pipeline in speaker verification challenges~\cite{voxsrc2022}.
\par
Most speaker SSL models have been based on training objectives proposed in the computer vision~(CV) community.
For example, the studies in~\cite{spkssl_simclr_moco,spkssl_vicreg} applied self-supervised contrastive learning approaches that maximize and minimize the similarity between output embeddings extracted from the same and different samples, respectively.
More recently, motivated by the success of the non-contrastive framework, there have been several proposals for speaker SSL models trained without negative samples that are different utterances from a given anchor~\cite{spkssl_vicreg,spkssl_dino_er,spkssl_c3dino,spkssl_dino_compara,spkssl_ca_dino,spkssl_rdino}.
Among non-contrastive frameworks, the \textit{self-DIstillation with NO labels} (DINO)~\cite{dino_org} based method has been a promising approach reporting outstanding performance in the ASV task under unsupervised conditions~\cite{spkssl_dino_er,spkssl_c3dino,spkssl_dino_compara,spkssl_ca_dino,spkssl_rdino}.
\par
On the basis of previous observations, we adopt a DINO-based speaker SSL model~\cite{spkssl_dino_compara}.
The model architecture follows ECAPA-TDNN but is trained using the DINO training scheme.
This non-contrastive method is based on a self-distillation framework that transfers the knowledge of a teacher model to a student model.
The teacher model is derived from an exponentially moving average of the student parameters, resulting in the same architecture but with different parameters.
To obtain a representation consistent with an utterance and invariant to variations of channel information~e.g.,~noise and reverberation, the inputs are multiple segments randomly extracted from an utterance, each of which is applied to different data augmentations.
During training, the similarity between the student and teacher output is measured by a cross-entropy loss modified by either or both of \textit{sharpening} and \textit{centering} manipulation to avoid a collapse issue~\cite{dino_org}.

\vspace{-0.2cm}
\subsection{Analysis methods}
\vspace{-0.2cm}
\noindent \textbf{SUPERB evaluation probing task:}
To study what speech factors are captured by various speech models, we employ a probing task based on SUPERB~\cite{superb} as illustrated in Fig.~\ref{fig:ecapa_s3prl}.
This benchmark contains a variety of speech-processing tasks and has been well-studied in previous works~\cite{wavlm,superb_right,ashi_deep_wide}.
SUPERB uses the weighted sum of the hidden feature of different blocks, $\mathbf{x}_l$, as output of the upstream model,~i.e.,~$\sum\nolimits_{l=1}^{L}w_l\cdot\mathbf{x}_l$, where $w_l$ are trainable weights, $l$ is the block index, and $L$ is the number of blocks.
Analyzing the weights, $w_l$, allows us to identify the contribution of each block to solving a specific task.
Note that the intermediate output from ECAPA-TDNN has a different number of dimensions across the blocks unlike the speech SSL models, and hence, we insert a trainable linear projection $\mathbf{A}_l$ before the weighted-sum manipulation: $\sum\nolimits_{l=1}^{L}w_l\cdot(\mathbf{A}_l\mathbf{x}_l)$.
Furthermore, since the last segment-level block outputs a single vector from an input sequence, we add the repeat operation.
To analyze the contribution of each layer when using the ECAPA-TDNN features, to account for the contribution of the linear transformation, we multiply the weights, $w_l$, with the Frobenius norm of the linear transformation matrix as $w_l\cdot\| \mathbf{A}_l \|_F$.

\noindent \textbf{Measurement of layer-wise similarity:}
Since the above probing approach compares the layer representations indirectly, we additionally adopt a more direct approach to unveil how their representations differ.
Specifically, we use a layer-wise similarity analysis based on linear centered kernel alignment~(LinCKA)~\cite{lincka}, which makes it possible to gain deep insight into the representation of deep neural networks in the various modalities~\cite{deep_vs_wide,sphssl_similarity,lm_lincka,asr_lincka}.
LinCKA is a normalized Hilbert-Schmidt independence criterion (HSIC) between two similarity matrices, each calculated from the features derived from two different layers.
Unlike the CV community, the output features of hidden layers are temporal sequences calculated even from a single sample, leading to severe memory consumption when computing the original LinCKA.
Therefore, we utilized an unbiased version of LinCKA in a minibatch fashion as proposed in~\cite{deep_vs_wide}.

\begin{table*}[ht]
\caption{Evaluation result for each model and each task on selected SUPERB tasks. \texttt{LS}, \texttt{LL}, \texttt{GS} and \texttt{VC} denote LibriSpeech, Libri-Light, GigaSpeech, VoxPopuli and VoxCeleb2 respectively.}
\label{tab:indirect_result}
\vspace{-0.3cm}
\begin{center}
\scalebox{0.7}[0.7]{
\begin{tabular}{l|l|c|l||c|c|c|c|c|c}
\toprule
Upstream & Upstream & Model & Pre-training & SID & ASV-IDT (ASV) & KS & PR & ER & IC \\ \cline{5-10}
group & model & size & data & Acc$\uparrow$ & EER$\downarrow$ & Acc$\uparrow$ & PER$\downarrow$ & Acc$\uparrow$ & Acc$\uparrow$ \\
\hline
\midrule
Baseline & FBANK & - & - & 0.1 & 13.7 (10.3) & 8.3 & 96.9 & 29.0 & 4.5 \\
\midrule
\multirow{5}{*}{Speech SSL model}  & HuBERT \textsc{Base} \cite{hubert} & 94M & \texttt{LS} & 66.6 & 4.9 (5.5) & 96.6 & 5.7 & 65.1 & 98.2 \\
 & HuBERT \textsc{Large} \cite{hubert} & 315M & \texttt{LL} & 89.4 & 4.3 (5.6) & 94.9 & 4.3 & 66.0 & 98.4 \\
 & WavLM \textsc{Base} \cite{wavlm} & 94M & \texttt{LS} & 62.2 & 5.1 (5.7) & 96.3 & 5.6 & 64.9 & 98.5 \\
 & WavLM \textsc{Base+} \cite{wavlm} & 94M & \texttt{LL} + \texttt{GS} + \texttt{VP} & 72.3 & 3.9 (5.2) & 97.5 & 4.4 & 68.4 & \textbf{98.8} \\
 & WavLM \textsc{Large} \cite{wavlm} & 315M & \texttt{LL} + \texttt{GS} + \texttt{VP} & 92.9 & 2.8 (5.1) & \textbf{98.0} & \textbf{4.3} & \textbf{69.3} & 98.7 \\
\midrule
\multirow{2}{*}{Speaker model} & DINO & 15M & \texttt{VC} & 97.6 & 2.8 (8.0) & 96.7 & 33.8 & 59.8 & 88.5 \\
 & Supervised & 15M & \texttt{VC} & \textbf{98.5} & \textbf{2.2 (4.8)} & 96.3 & 34.1 & 60.4 & 84.1 \\
\bottomrule
\end{tabular}
}
\end{center}
\vspace{-0.9cm}
\end{table*}

\vspace{-0.2cm}
\section{Experimental setup}
\vspace{-0.3cm}
Our implementation was based on WeSpeaker\footnote{\url{https://github.com/wenet-e2e/wespeaker}}~\cite{wespeaker} for training the speaker models and S3PRL\footnote{\url{https://github.com/s3prl/s3prl}}~\cite{superb} for benchmarking on SUPERB.
\vspace{-0.2cm}
\subsection{Speaker models}
\label{ssec:setup_spk_models}
\vspace{-0.2cm}
To train the speaker SSL and supervised models, we utilized the development set of VoxCeleb2~\cite{voxceleb} according to the VoxSRC challenge~\cite{voxsrc2019}.
This split is one of the most widely used for text-independent speaker-related tasks and consists of almost 2300 hours of data spoken by 5994 speakers.
When feeding speech data to the models, 80-dimensional log mel-filterbank outputs were extracted with a 25-ms window and a 10-ms shift followed by a segment-wise cepstral mean normalization, called FBANK.
\par
The model architecture was ECAPA-TDNN, the same as the original paper~\cite{ecapa_tdnn} consisting of 1024 channels of convolution layers in the frame-level blocks~(the standard 1D convolution layer, multiple SE-Res2Blocks).
\par
The rest of the settings for training both speaker SSL and supervised models were identical to the configurations of each VoxCeleb recipe\footnote{\texttt{voxceleb/v2} for supervised model and \texttt{voxceleb/v3/dino} for speaker SSL} on the WeSpeaker toolkit.

\vspace{-0.1cm}
\subsection{Speech SSL models}
\vspace{-0.2cm}
For comparison with the speaker models, we utilized publicly available speech SSL models.
Specifically, we employed the \textsc{Base} and \textsc{Large} versions of the HuBERT~\cite{hubert} and WavLM~\cite{wavlm} models.
The \textsc{Base} and \textsc{Large} architectures consist of 12 and 24 Transformer blocks, respectively, on top of 7 convolutional blocks.
The training data for all \textsc{Base} models was the LibriSpeech dataset.
HuBERT~\textsc{Large} was trained on only Libri-Light and WavLM~\textsc{Large} on a combination of Libri-Light, GigaSpeech, and VoxPopuli.
We also evaluated the WavLM~\textsc{Base+} model, which has the same structure as WavLM~\textsc{Base} but is trained on the same data as WavLM~\textsc{Large}.

\vspace{-0.2cm}
\subsection{Comparison settings}
\vspace{-0.2cm}
For probing various models on SUPERB, we selected six tasks from each category,~i.e.,~the content, speaker, semantic, and paralinguistic categories suggested in \cite{superb}.
From the content category, we employed the phoneme recognition (PR) and keyword spotting (KS) tasks.
As for the speaker category, we experimented with SID and ASV tasks.
In addition to the original ASV task that trains the TDNN downstream model, we investigated a simpler downstream model.
Specifically, we substituted the original model with a parameter-less module that directly forwards the input.
This modified ASV task with identity processing is referred to as ASV-IDT.
Since the speaker models had already been trained on the ECAPA-TDNN architecture and acquired sophisticated speaker representation, this TDNN downstream model would have been redundant and had the possibility of degrading the performance of the ASV task.
Moreover, since the recent study of~\cite{superb_right} also reported that the downstream architecture affects task performance significantly, it is essential to investigate different downstream models.
Regarding the semantic and paralinguistic categories, we executed intent classification (IC) and emotion recognition (ER) tasks, respectively.
For the comparison condition, we used the FBANK upstream explained in Section~\ref{ssec:setup_spk_models} as a baseline in addition to the speech SSL, speaker SSL, and supervised speaker models.
\par
For the similarity analysis, we used a subset of VoxCeleb1, which extracted 4 samples from each speaker's utterances, resulting in a total of 5004 samples.
To calculate the minibatch LinCKA, we adopted a batch size of 4.
When comparing ECAPA-TDNN with speech SSL models, since the temporal resolution of ECAPA-TDNN is almost twice as high as speech SSL models, we repeated each element of the speech SSL output twice to match its sequence length.

\vspace{-0.2cm}
\section{Results}
\label{sec:exp_res}
\vspace{-0.3cm}

\subsection{SUPERB evaluation probing tasks}
\label{ssec:exp_res_superb}
\vspace{-0.2cm}
\noindent \textbf{Performance comparison:}
Table~\ref{tab:indirect_result} shows the performance results from evaluating each upstream model on a subset of the SUPERB tasks.
Among the speech SSL models, the WavLM~\textsc{Large} model achieved the best performance across nearly all tasks.
However, the supervised and DINO speaker models outperformed WavLM in the SID and the ASV tasks, suggesting the acquisition of a more sophisticated speaker representation.
This tendency is somewhat reasonable because the speaker models were trained with speaker-specialized criteria, unlike the speech SSL models trained for general-purpose representations.
Interestingly, both speaker models can solve the non-speaker tasks significantly better than the baseline model, indicating that various information was obtained even in the supervised speaker model.
When comparing the two speaker models, the supervised model achieved better performances in the speaker tasks trivially but worse ones in the non-speaker tasks except for the ER task.
These results imply that simply reinforcing non-speaker representations does not necessarily improve the accuracy of speaker tasks.

\begin{figure}[t]
  \centering
  \includegraphics[width=5cm]{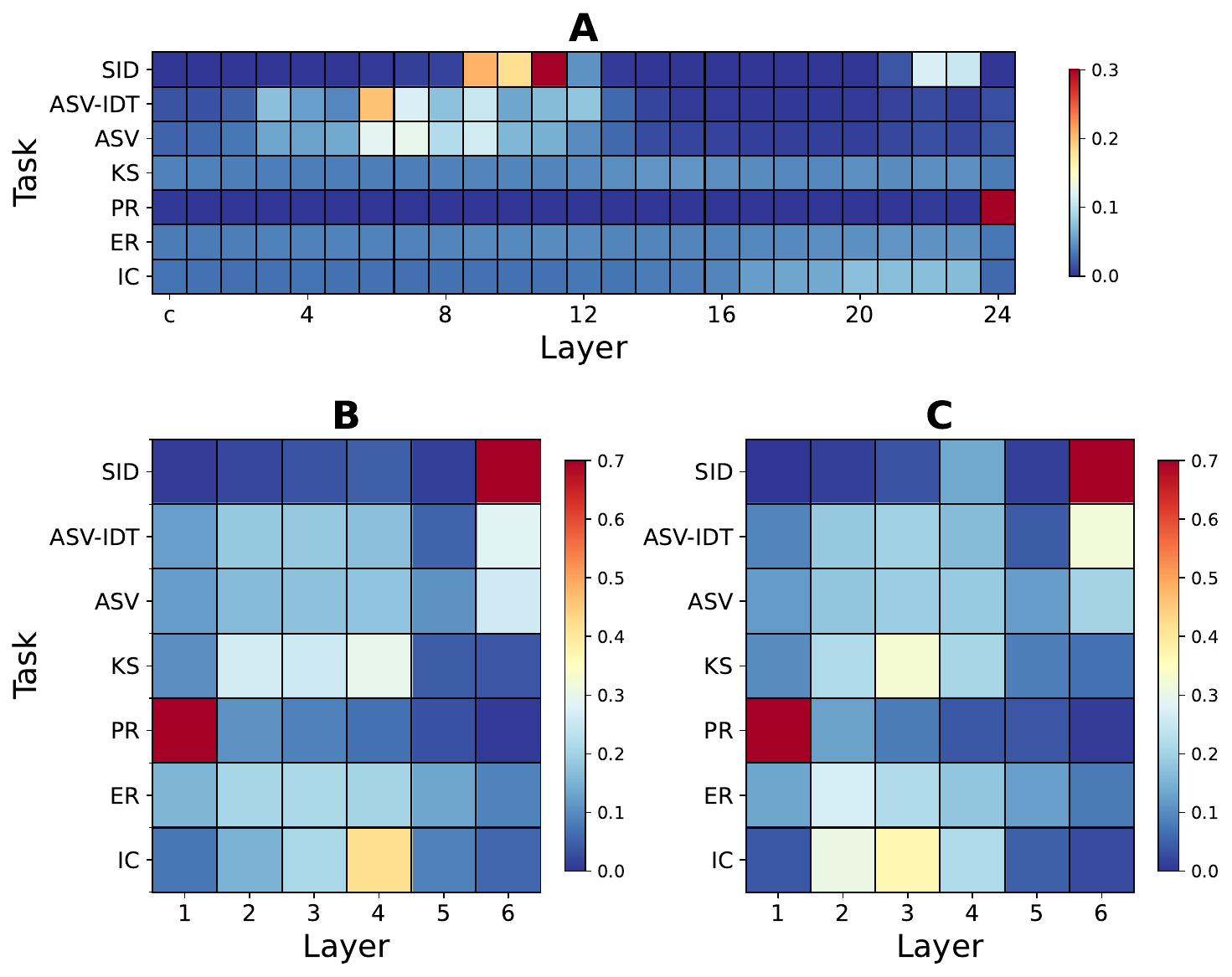}
  \vspace{-0.4cm}
  \caption{Visualization results of weights of weighted sum for (A) WavLM \textsc{Large}, (B) DINO, and (C) supervised speaker model.}
  \vspace{-0.5cm}
  \label{fig:weights_sum}
\end{figure}

\noindent \textbf{Weight comparison of weighted sum:}
Fig.~\ref{fig:weights_sum} shows the weights of the weighted sum for each model on each task.
Note that the ``c'' layer in Fig.~\ref{fig:weights_sum}A corresponds to the input to the 1st block of the Transformer encoders.
As seen in Fig.~\ref{fig:weights_sum}A, the WavLM \textsc{Large} model captured a speaker representation mainly in the lower layers (up to layer 12) and phoneme and semantic information in the latter layers as presented in~\cite{wavlm}.
Figs.~\ref{fig:weights_sum}B and C show that for the two speaker models, similar weight tendencies were observed in the ASV, PR, and ER tasks.
In particular, the PR task focused on the 1st frame-level block, indicating that when optimizing on the basis of speaker criteria, the hierarchical order in which speech factors are captured would be the opposite of WavLM.
On the other hand, the SID, KS, and IC tasks seemingly exploited somewhat different layers between the two speaker models.
Concretely, the 4th layer of the supervised speaker model captured more information related to SID and less related to KS and IC than the speaker SSL model.
The highest weight for PR on the first layer, combined with the relatively poor performance on the PR probing task as shown in Table~\ref{tab:indirect_result}, suggests that powerful speaker representations for SID tend to disentangle phoneme information rapidly.
Moreover, the representations related to IC and KS in the speaker SSL model appeared in later layers compared with the supervised model, suggesting that capturing spoken content information in layers closer to the input tends to be more suitable for the speaker tasks.
\vspace{-0.1cm}
\subsection{Layer-wise similarity comparison}
\vspace{-0.1cm}
In this section, we quantify the similarity of layer representations through LinCKA.
Note that, in the subsequent similarity figures, bright colors indicate high similarity between two layers.
\par
\begin{figure}[t]
  \centering
  \includegraphics[width=8cm]{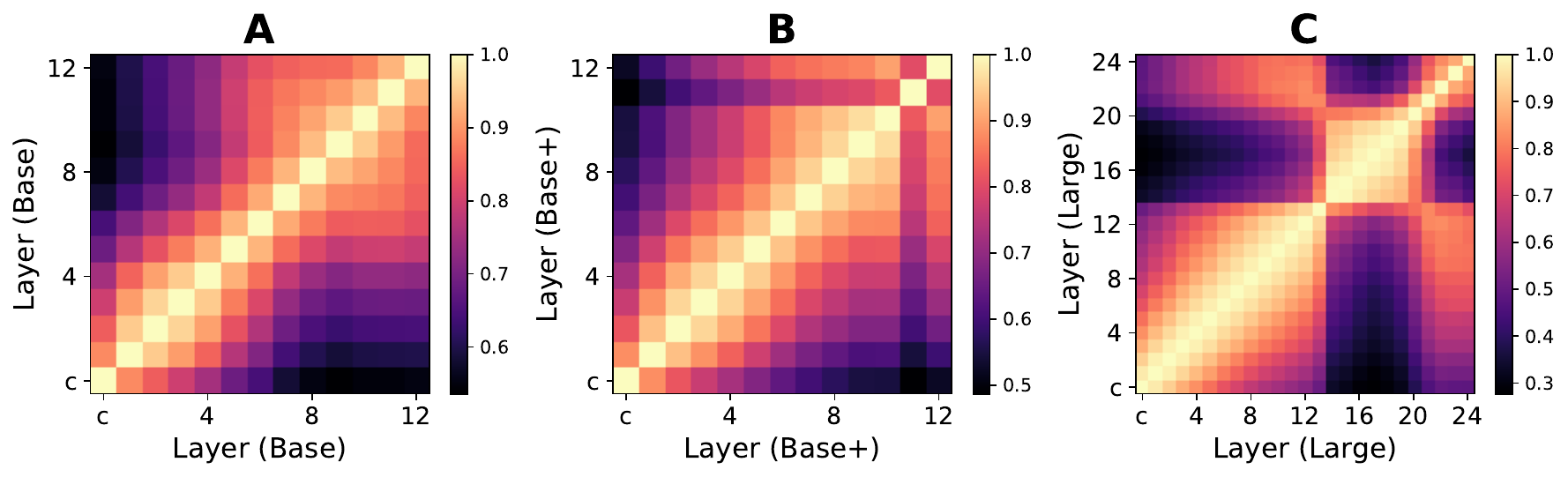}
  \vspace{-0.4cm}
  \caption{Similarity results of each layer comparing identical WavLM models with variation of (A) \textsc{Base}, (B) \textsc{Base+}, and (C) \textsc{Large}.}
  \label{fig:wavlm_sanity}
  \vspace{-0.3cm}
\end{figure}

First, we examine the representation between layers of identical models.
Fig.~\ref{fig:wavlm_sanity} shows the similarity between the layers of the same WavLM \textsc{Base}, \textsc{Base+} and \textsc{Large}.
High similarity around the diagonal of Fig.~\ref{fig:wavlm_sanity}A clearly confirms that adjacent layers were more similar.
On the other hand, Figs.~\ref{fig:wavlm_sanity}B and C show distinctive representations around layer 11 for \textsc{Base+} and layers 13 to 22 for \textsc{Large}, resulting in an exceptionally low similarity to other layers close to the input and output.
For WavLM~\text{Large}, some previous work~\cite{sphssl_lwise_anal,speechglue} indicated that these intermediate layers partly seemed to capture linguistic information such as word, semantic, and syntax, leading to acquiring a different representation from the layers processing acoustical information~(e.g.,~speaker and phonetic information).
This consideration is also consistent with the result that speaker and phonetic information are distributed in either or both early and late layers as shown in Fig.~\ref{fig:weights_sum}A.
Note that the difference between WavLM~\textsc{Base} and \textsc{Base+} is only the amount of unlabeled data~(i.e., 960 and 94k hours), which suggests that this representation of information would arise only when using a sufficient amount of training data.
\par
\begin{figure}[t]
  \centering
  \includegraphics[width=5.5cm]{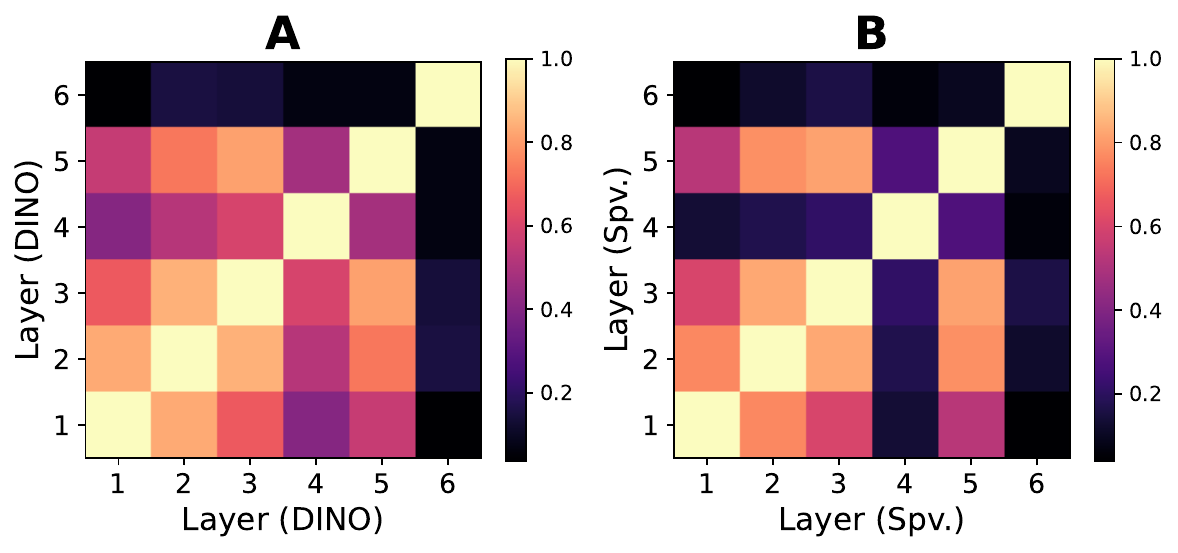}
  \vspace{-0.4cm}
  \caption{Similarity results of each layer comparing identical (A) DINO and (B) supervised speaker model.}
  \label{fig:spk_sanity}
  \vspace{-0.4cm}
\end{figure}

Figs.~\ref{fig:spk_sanity}A and B demonstrate heatmaps for the speaker SSL model and the supervised speaker model, respectively.
Both heatmaps show characteristic representations, which differ from the other layers, in the 4th and 6th layers corresponding to the final SE-Res2Block and the speaker embedding layer.
\par
\begin{figure}[t]
  \centering
  \includegraphics[width=7cm]{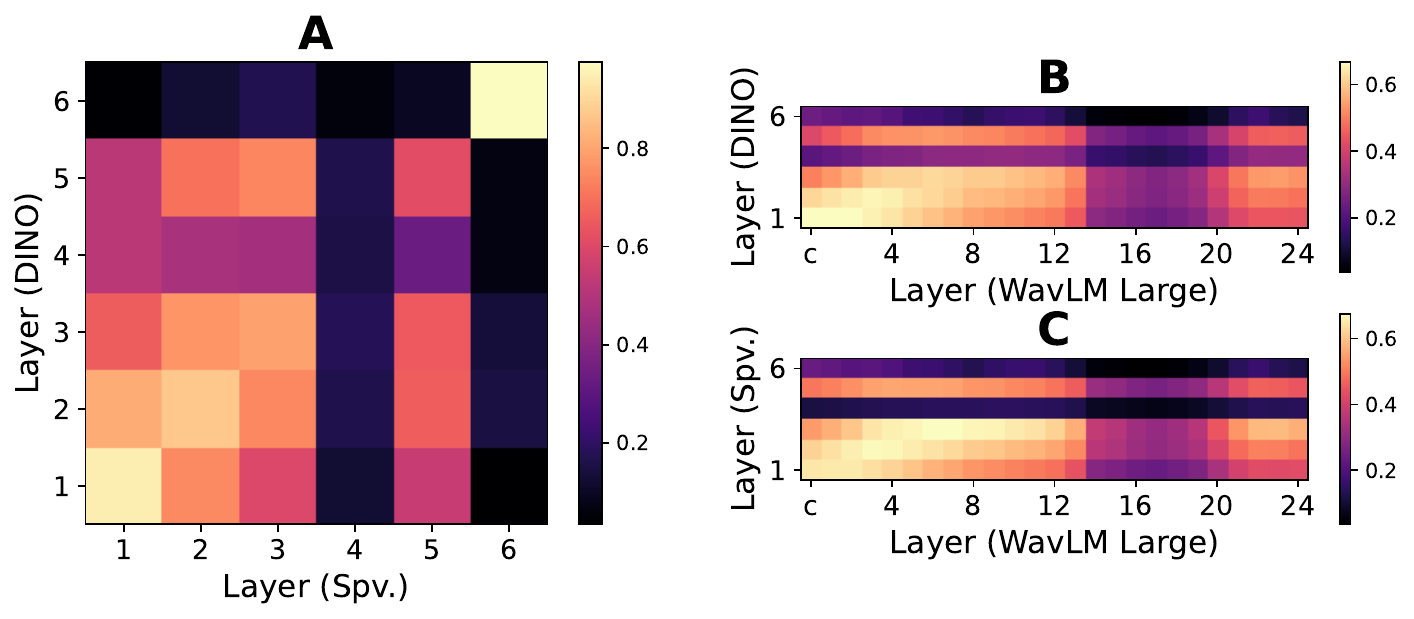}
  \vspace{-0.4cm}
  \caption{Similarity results of each layer comparing (A) DINO with supervised speaker model, (B) DINO with WavLM \textsc{Large}, and (C) supervised speaker model with WavLM \textsc{Large}.}
  \label{fig:dino_spv_wavlm_similarity}
  \vspace{-0.2cm}
\end{figure}

Fig.~\ref{fig:dino_spv_wavlm_similarity} shows the similarity between layers of different models.
In Fig.~\ref{fig:dino_spv_wavlm_similarity}A, a comparison between the speaker SSL and supervised speaker models shows that while all conditions of the two models were identical except for the training criterion, relatively low similarities are observed between layers except between the first and the last layers.
This result indicates that both final speaker representations are similar, but their extraction processes are somewhat different.
In particular, the representation of the 4th layer in the supervised model exhibited significant differences compared with any of the layers in the speaker SSL model.
Conversely, the representation of the same layer in the speaker SSL model appears to be distributed across multiple layers in the supervised speaker model.
The unique representation of the 4th layer in the supervised speaker model seems to provide auxiliary information for the SID task when comparing the 4th layers of Figs.~\ref{fig:weights_sum}B and C.
This may be a reason explaining the better performance of the supervised speaker models compared with the SSL one for the SID task.
Additionally, from Figs.~\ref{fig:weights_sum}B and C, we observe that the speaker SSL model often used the 4th layer to solve KS and IC tasks, where it achieved higher performance than the supervised speaker model.
This suggests that the speaker SSL model has a lower capacity to represent speaker information because it captures more other speech-related information.
Note that the distinct representation in the 4th layer of the two speaker models is also dissimilar from the WavLM model as shown in Figs.~\ref{fig:dino_spv_wavlm_similarity}B and C.
More diverse probing tasks may be necessary to clarify what information is captured by the 4th layer of the two speaker models, such as channel and speaking style classification.
\par
From Figs.~\ref{fig:dino_spv_wavlm_similarity}B and C, we also find a low similarity between the layers of the speaker models and WavLM in the later layers,~i.e.,~layers 13 to 22.
This indicates that the speaker models cannot capture linguistic information well.
Furthermore, from Figs.~\ref{fig:dino_spv_wavlm_similarity}B and C, the 1st layer of the speaker models, which is important for solving the PR task, is most similar to the early layers of WavLM.
This result suggests that the phoneme representation of the speaker models is significantly rough compared with WavLM and that there is room for improvement by reinforcing the capacity of phoneme identity~\cite{tawara_san}.
\par
\begin{figure}[t]
  \centering
  \includegraphics[width=8cm]{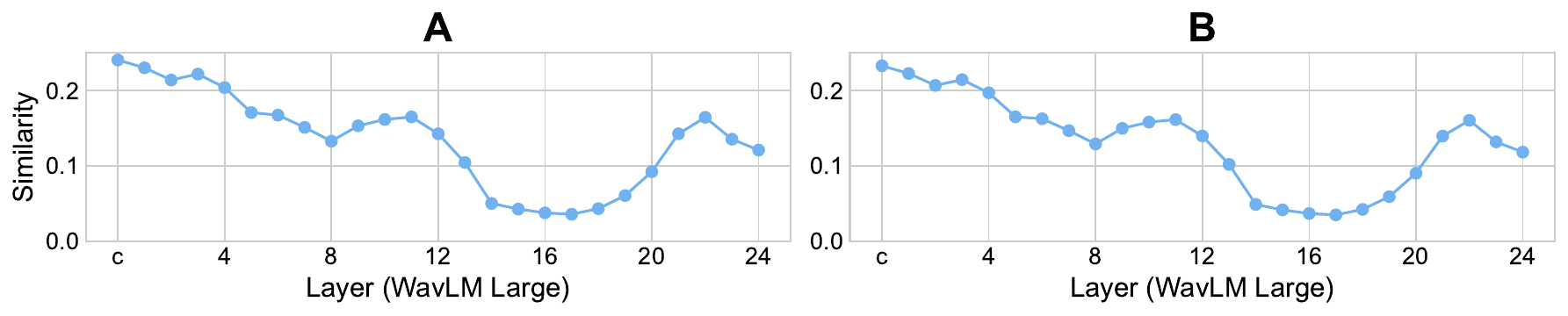}
  \vspace{-0.4cm}
  \caption{Similarity results between each layer of WavLM \textsc{Large} and last layer of (A) DINO and (B) supervised speaker model.}
  \label{fig:wavlm_spkinfo}
  \vspace{-0.4cm}
\end{figure}

Finally, to analyze the relationship between the speaker embeddings and WavLM, Fig.~\ref{fig:wavlm_spkinfo} shows the similarity between the final layer of the speaker models and each layer of WavLM \textsc{Large}.
The WavLM model included speaker information not only in the early layers but also in the 21--23rd layers.
This result is consistent with the result from Fig.~\ref{fig:weights_sum}A, which shows the high contribution of a similar layer in the SID task.


\vspace{-0.2cm}
\section{Conclusion}
\label{sec:con}
\vspace{-0.2cm}
In this work, we endeavored to unveil how speech SSL, speaker SSL and fully-supervised speaker models represent information and how they differed from each other through probing tasks and similarity analysis.
Our experiments suggest that the two speaker models exhibit distinct representations in intermediate layers, leading to performance gaps between them.
Furthermore, it appears that these models have not achieved a profound level of linguistic representation.
We hope that these results will help in better understanding these models and lead to the development of more efficient models and learning schemes.

\clearpage
\bibliographystyle{IEEEbib}
\bibliography{refs}
\end{document}